\documentclass{aadebug}
\usepackage{a4wide}
\usepackage{latexsym}
\usepackage{alltt}
\usepackage{algorithmic}
\usepackage{algorithm}
\usepackage{german}

\corr{0309027}{197}
\begin{document}

\runningheads{Bernhard Peischl and Franz Wotawa}{Modeling State in
Software Debugging of VHDL-RTL Designs -- A Model-Based Diagnosis
Approach}

\title{Modeling State in Software Debugging of VHDL-RTL Designs -- A Model-Based Diagnosis Approach}

\author{Bernhard Peischl\addressnum{1} and Franz Wotawa\addressnum{1},\extranum{1}}

\address{1}{
  Technische Universit"at Graz,
  Institute for Software Technology (IST),
  Inffeldgasse 16b/2, A-8010 Graz, Austria,
  \tt\{peischl,wotawa\}@ist.tu-graz.ac.at}

\extra{1}{Authors are listed in
        alphabetical order.}





\begin{abstract}
In this paper we outline an approach of applying model-based diagnosis to the
field of automatic software debugging of hardware designs. We present our value-level
model for debugging VHDL-RTL designs and show how to localize the erroneous
component responsible for an observed misbehavior. Furthermore, we
discuss an extension of our model that supports the debugging of sequential
circuits, not only at a given point in time, but also allows for considering
the temporal behavior of VHDL-RTL designs. The introduced model is capable of
handling state inherently present in every sequential circuit. 
The principal applicability of the new model
is outlined briefly and we use industrial-sized real world examples from the
ISCAS'85 benchmark suite to discuss the scalability of our approach.  
\end{abstract}

\section{Introduction}

During the last decade, hardware description languages like Verilog and
VHDL (Very High Speed Integrated Circuit Hardware Description Language) have
become common for designing circuits. In order to reduce time to market, which becomes increasingly
important in today's fast paced economy, applying formal verification
techniques and extensive simulation is imperative in order to master
the error-prone development process of complex circuit designs. However, if some
faulty behavior is identified due to simulation or automated verification, the
location of the error in the source code is of particular
interest. Nevertheless, there is currently almost no support for the designer in
fixing the faults thus found. Traditional debuggers force the developer to go
through the code and therefore fault localization and error
correction is a time consuming task.

To overcome these issues several approaches for automated fault
localization have been proposed including program-slicing \cite{wei82,wei84},
algorithmic debugging \cite{sha83}, dependency-based techniques
\cite{kup89,korel88,jackson95} and probability-based methods
\cite{burhor93}.
Moreover, in the recent years, the usage of model-based diagnosis \cite{rei87,kle87,gre89} for debugging
software has been examined in a wider context \cite{fsw99,msw99,sw99a}. All of these approaches
have in common that they use a model that can be
automatically derived from the source code, they use different abstractions
and also differ in the
programming languages being considered. The spectrum ranges from
purely logical languages to imperative languages and hardware description
languages such as VHDL.

In this paper we focus on software debugging of hardware designs, particularly
on our modeling approach for VHDL programs. 
By using a domain-independent diagnosis engine together with a specific
model abstracting syntax and semantics of the programming language,
this approach allows for automated debugging of logical, imperative \cite{fsw99,msw99}
and functional programming languages \cite{sw99a} . By changing the underlying model
the approach can be tailored to specific programming languages
without the need for changing the underlying algorithm for fault localization.  

In this paper we present an extension to previous research \cite{wot00b,wot00d}
on debugging VHDL designs by introducing a model extension that allows not only for fault
localization at a given point in time but also allows for considering
the temporal behavior of a VHDL program. The new model improves the
previous model in that it explicitly captures process semantics by unfolding
process executions over time and allows for representing the state of
sequential circuit designs. This is done by  unfolding the model in time and by
creating temporal instances of components for every possible process execution.

The remainder of the paper is organized as follows. The next section outlines
the main ideas behind model-based diagnosis. Section \ref{debugging_vhdl}
applies model-based diagnosis to localize faults in VHDL programs by presenting a
small but realistic example. We use a 2-bit counter to show our
modeling approach for a single point in time. Section \ref{temporal_abstraction} introduces our
new model extension that allows for considering the temporal behavior of
the design rather than using a single point in time and Section \ref{empirical_results} outlines some empirical results
obtained from real-world examples.  An overview summarizing the paper's
main contribution and a discussion on related and forthcoming research concludes the paper. 

\subsection{Model-Based Diagnosis and Software Debugging}

The basic idea of model-based diagnosis is to use
knowledge of the correct behavior of the components of a system together with
knowledge of the structure of the system to locate the cause of the
misbehavior. The behavior and the structure of the system are the model and
the components of the system are the parts that behave either correct or
abnormal. 

A model-based diagnosis engine assumes that some components are faulty
and the remaining ones are not. This approach requires a specification
of the correct behavior of the components but it does not require 
knowledge about the behavior of a faulty component. The diagnosis engine
checks whether the model contradicts some given observations under certain
assumptions about the abnormality of components. If there is no contradiction,
then the assumption is justified and the assumed faulty components are a valid diagnosis of the
given system.

The crucial decision when building a diagnosis model from program code
involves the choice of those parts of the program which are 
represented as components in the model. This decision directly influences the
granularity of the diagnosis results. For example, if
statements are mapped to components, bugs can only be localized on the
statement level. Faults inside a statement cannot be diagnosed without
fault models, that is, knowledge about how a particular type of fault may
change the system's behavior. Using expressions as diagnosis
components allows finer granularity at the cost of increasing the number of
components and thus the overall diagnosis time increases considerably.



Using the consistency-based view on diagnosis as defined by \cite{rei87}, a
diagnosis system can be formally stated as a tuple $(SD,COMP)$ where $SD$
(system description) is a
logical theory modeling the behavior of the given program to be debugged, and
$COMP$ is a set of components, i.e., statements or expressions. A diagnosis
system together with a set of observations $OBS$, that is, a test case, forms
a diagnosis problem. A diagnosis $\Delta$ is a subset of $COMP$, with the
property that the assumption that all components in $\Delta$ are
abnormal (they do not behave as expected), and the rest of the components is behaving normal
(they behave as expected), is
required to be consistent with $SD$ and $OBS$. Formally, $\Delta$ is a
diagnosis if and only if $SD \cup OBS \cup \{\neg AB(C)|C \in COMP \setminus \Delta\}
\cup \{AB(C)|C \in \Delta\}$ is consistent, where $\neg AB(C)$ formalizes a
correctly operating component. Note, that in contrast to correctness, which in
the terminology of software engineering requires that there is no
contradiction taking into account all test cases, the notion of consistency
in the context of model-based diagnosis refers to
logical validity with respect to a given test case under a 
given set of assumptions. 

The basis for localizing faults is that an incorrect output value (where the incorrectness
can be observed directly or derived from observations of other signals) cannot
be produced by a correctly functioning component with correct
inputs. Therefore, to make a system with observed incorrect behavior
consistent with the description, some subset of its
components has to be assumed to work incorrectly. In practical terms, one is
interested in finding minimal diagnosis, i.e., a minimal set of components
whose malfunction explains the misbehavior of the system. 

In principle all diagnoses can be computed by successively testing all subsets
of $COMP$ . However, in automated debugging this approach can only be applied for rather small
programs and is not feasible for industrial sized designs. 
Reiter \cite{rei87} proposed the use conflicts. A set $CO \subseteq COMP$
is a conflict if and only if $SD \cup OBS \cup \{\neg AB(C)|C \in CO\}$ is
contradictory, i.e., assuming that the statements in $CO$ are abnormal 
leads to inconsistency. 
The relationship between diagnosis and
conflicts can be explained as follows. A conflict states that $\neg AB(C_1)
\wedge \neg AB(C_2) \wedge ... \wedge \neg AB(C_n)$ contradicts the
observation. In order to resolve the contradiction at least one of the
assumptions $\neg AB(C_1), ... \neg AB(C_n)$ has to be false, i.e., there must
be a $C_i$; $ 0 \le i \le n$ where $AB(C_i)$ is true. If we want to resolve
all conflicts we have to choose one (not necessarily different) element from
every conflict. Since the so chosen elements resolve all conflicts, they must
form a diagnosis. 

A set with the property that the intersection with all
elements of a set of sets is not empty is called a hitting set. Hence, a
minimal diagnosis (a diagnosis where no subset is a diagnosis) is a minimal hitting set of (not necessarily minimal)
conflicts. From this observation follows that there exists only a single-
diagnosis (a diagnosis with exactly one component) if and only if the
component is element of every conflict. Reiter \cite{rei87} proposed an
algorithm to compute all minimal diagnoses from the conflicts by means of a so
called hitting set tree. There are algorithms where this process is done in an iterative fashion,
that is, conflicts are computed on demand. Although computing all
conflicts is of order $2^{|COMP|}$ the computation of single conflicts (which
are not minimal) can be done in linear time when assuming the the model is
written as a set of propositional Horn clauses. Alternatively, there are other
very fast diagnosis algorithms that rely on other principles \cite{sw97b,dechter95}. 

\section{Debugging of VHDL-RTL-Programs}

\label{debugging_vhdl}

In the domain of debugging hardware designs the model is given by the syntax
of the program and the semantics of the language, which in our case is the RTL subset
(Register Transfer Level) of VHDL. Depending on the required granularity, components refer
to statements or expressions. In the following we present an introductory
example in which we assume that statements correspond to diagnosis
components. Hence, statements are the smallest parts of the program that can
be either correct of faulty. 

Figure \ref{counter_vhdl} outlines the VHDL code of a
2 bit counter that decrements its value if the input $E1$ is set to 1 and $E2$
is set to 0; if both inputs are set to 0 the counter increments its value
whenever a rising edge occurs on the $CLK$ signal. If $E1$ is set to 0
and $E2$ is set to 1  the counter is reset to 0, and finally, if both inputs are 1 the
counter is set to value 3. The output of the counter is coded using a 1-from-4
encoder. The value 0 is thus represented by $A1A2A3A4=1000$ whereas the value
3 is coded by $A1A2A3A4=0001$.

\begin{figure}
\centering{
\input{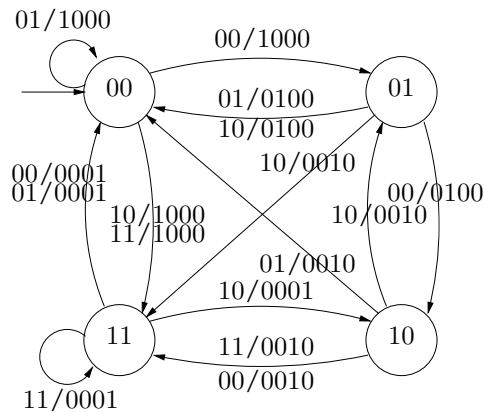}
}
\caption{The state-transition diagram.}
\label{state_transition_correct}
\end{figure}

The behavior of our program can be compiled into a simple state-transition
diagram, outlined in Figure \ref{state_transition_correct}.
There the values of the
signals $Q1$ and $Q2$ are given in circles (as states of the automaton), and
the inputs and corresponding outputs are given on the arcs (encoded as
$E1E2/A1A2A3A4$). The initial state is indicated by an arrow. 
\begin{figure}
\centering{
\scriptsize{
\begin{alltt}
entity COUNTER is
  port(
    E1, E2, CLK    : in  bit;
    A1, A2, A3, A4 : out bit);
   end COUNTER ;

    architecture BEHAV of COUNTER is
      signal D1, D2, Q1, Q2, NQ1, NQ2 : bit := 0;
    begin
       comb_in : process (Q1, Q2, E1, E2)
         variable I1, I2 : bit;
       begin
         I1 := not((Q1 and Q2) or (not(Q1) and not(Q2)));
         I2 := (I1 and E1) or (not(I1) and not(E1));
         D1 <= (E1 and E2) or (E2 nor I2);
         D2 <= (E1 and E2) or (E2 nor Q2);
       end process comb_in;

       comb_out : process (Q1, Q2, NQ1, NQ2)
       begin
          A1 <= NQ2 and NQ1;
          A2 <= Q2 and NQ1;
          A3 <= NQ2 and Q1;
          A4 <= Q1 and Q2;
       end process comb_out;

        
        p1:process (Q1)
        begin
                NQ1 <=  not(Q1);
        end process p1;

        p2:process(Q2)
        begin
                NQ2 <= not(Q2);
        end process p2;

       dff : process (CLK)
       begin
         if (CLK = '1') then
           Q1 <= D1;
           Q2 <= D2;
         end if;
       end process dff;
    end BEHAV;
\end{alltt}
}
}
\caption{The source code of the 2-bit counter.}
\label{counter_vhdl}
\end{figure}
\normalsize{
In order to state the idea of applying model-based diagnosis to locate the
cause of an observed misbehavior we introduce a bug into the program by simply
omitting the not operator in line 30:
}

\scriptsize{
    p1:process (Q1)
        begin
                NQ1 <=  Q1;
        end process p1;

\begin{figure}
\begin{center}
\input{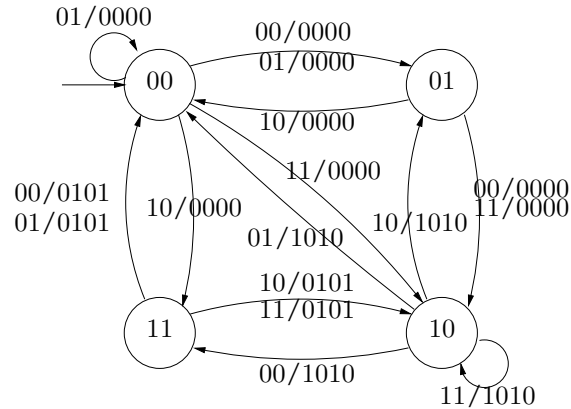}
\end{center}
\caption{The state-transition diagram of the faulty program.}
\label{state_transition_q1}
\end{figure}
}

\normalsize{
The state transition diagram of the erroneous program is given in Figure \ref{state_transition_q1}.
Considering this diagram, it is obvious that our design contains a bug: Since the
counter uses a 1 from 4 coded output, exactly one of the outputs should be 1
at any time, and this is not the case. From the failed property we can
identify the temporal context and retrieve the relevant state of the observed
misbehavior. Note that this can be done either manually by an experienced
designer or by means of automated verification techniques such as model
checking. By using the values of the retrieved state, the 
input signals and the erroneous output signals at a specific point in time as
observations we can use
model-based diagnosis to search for faulty statements that explain
the observed misbehavior.
}

\begin{figure}
\begin{center}
\input{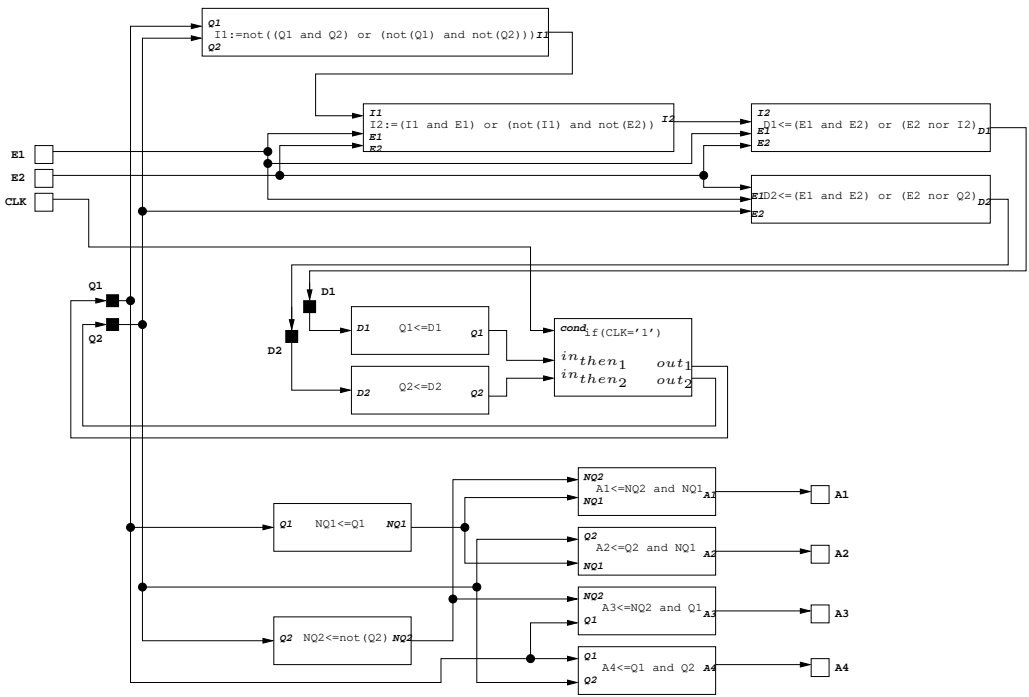}
\end{center}
\caption{The structural model of the counter.}
\label{model_struct}
\end{figure}

In order to locate the bug in the program, we have to build a model of the
program which comprises the structure and the behavior of the counter. The
structural part can be obtained by viewing statements as diagnosis components
and signals and variables as connections between them. For VHDL some semantic
aspects have to be considered, e.g., the different handling of variables and
signals. However, for sake of brevity we do not discuss these issues in
this article. Details can be found in \cite{wot00b,wot00d}. Figure \ref{model_struct}
shows the structural part of the model where components (statements in our
case) are indicated using boxes. 

The description of the components' behavior
remains to be introduced. The component model itself can be derived directly
from the associated statements. For example, the correct behavior of the
if-then-else statement can be represented by logical sentences. 
The output of an if-then-else statement is determined by so called
sub-blocks, that represent the statements in the if and the else
branch, respectively. If the component works as expected, that is,
$\neg AB(IF)$ is assumed to be true, and the condition evaluates to
true, the then sub-block, associated with the statements in the
if branch, is transferred to the output. Otherwise the evaluated
sub-block representing the else branch is assigned to the output. In
formal terms this can be expressed as follows:

$$ \forall x \in stmts_{then} \cdot \neg AB(IF) \wedge (cond(IF) = true)
\rightarrow (out_{x}(IF) = in_{then}(IF)) $$
$$\forall x \in stmts_{else} \cdot \neg AB(IF) \wedge (cond(IF) = false)
\rightarrow (out_{x}(IF) = in_{else}(IF))$$

In the formal model outlined above $stmts_{then}$ denotes the signal assignment-statements in the
if-branch, $stmts_{else}$ represents the statements in the
else-branch and the signal $out_{x}$ denotes a single output $x$.
Similar models can be introduced for all the other components in the
example. Details of modeling other VHDL-RTL language artefacts 
can for example be found in \cite{wot00d}.

For locating the
fault in our program automatically, we make use of the available knowledge
about the values of the signals. For our example we consider the detected
misbehavior for the case of counting downwards ($E1=1$, $E2=0$). According to
the state-transition diagram in Figure \ref{state_transition_q1} the
misbehavior is encountered several times. However, retrieving an appropriate state for
computing diagnoses is in the responsibility of the designer. From the failed property we know that the
program behaves wrong if $CLK=1$, $E1=1$, $E2=0$, $Q1=1$ and
$Q2=0$. We now take the expected output values and the values from
the state-transition diagram of the erroneous program to identify the
conflicts. Note that the values of $Q1$ and $Q2$ are known for two successive
states, hence we are able to eliminate the circular structure in Figure \ref{model_struct}. 

\begin{figure}
\begin{center}
\input{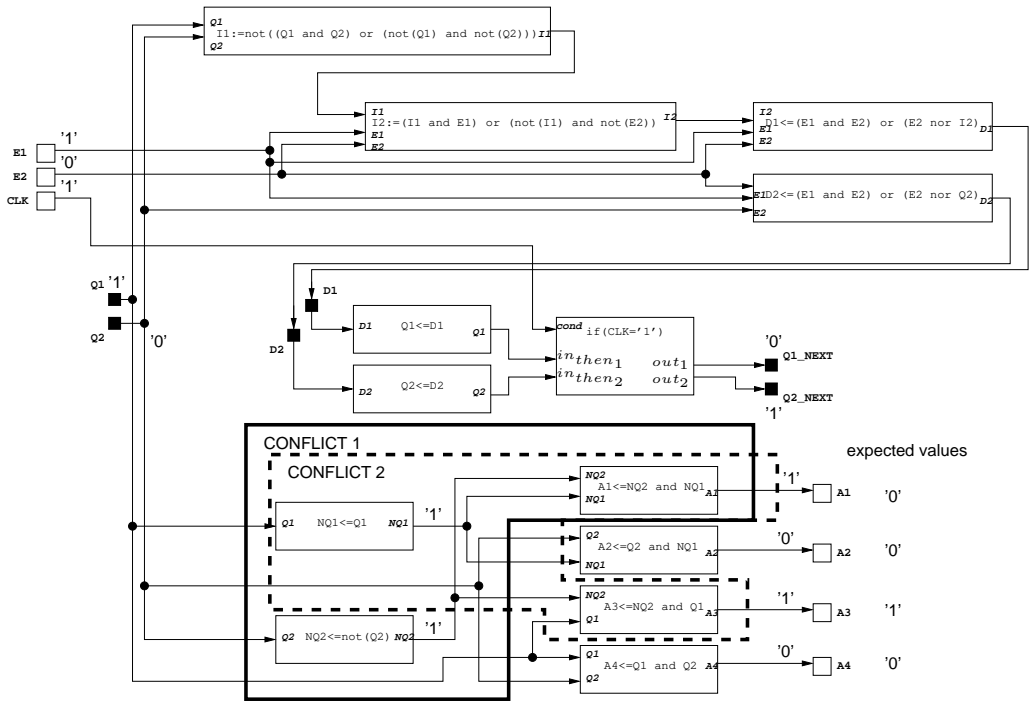}
\end{center}
\caption{The computation of the conflicts in the counter model.}
\label{conflict}
\end{figure}

According to
the transition-diagram of the correct program, we expect
the values $A1A2A3A4=0010$ rather than the produced output
$A1A2A3A4=1010$ for the given input signals. Thus, recalling the definition of conflicts, we obtain two
conflicts: One conflict comprises the statements $\{NQ1<=Q1$, $NQ2<=not(Q2)$,
$A1<=NQ2$ $and$ $NQ1$\}, and the other one consists of $\{NQ1<=Q1$, $A3<=NQ2$
$and$ $Q1$,
$A1<=NQ2$ and $NQ1$\}. Figure \ref{conflict} outlines the computation of the
conflicts graphically. 

Using these conflicts we can compute diagnoses by employing a
hitting set tree. In general, we have to built a directed acyclic graph \cite{gre89}, but
in our special case we simply obtain a tree.
 Our hitting set tree, called $T$ in the following, is a smallest
edge-labeled and node-labeled tree with the following properties \cite{rei87,gre89}:

\begin{itemize}
\item{The root of the tree is labeled by an arbitrary conflict.}
\item{For each node $n$ of $T$, let $H(n)$ denote the set of edge-labels on the
    path in $T$ from the root node to $n$. The label for $n$ is any conflict
    $CO$ such that $CO \cap H(n) = \{\}$, if such a conflict exists. Otherwise
    the label for node $n$ is $\surd$. If $n$ is labeled by the set $CO$, then
    for each $\sigma \in CO$, $n$ has a successor, $n_{\sigma}$, joined to $n$
    by an edge labeled by $\sigma$.}
\end{itemize}
 
\begin{figure}
\begin{center}
\input{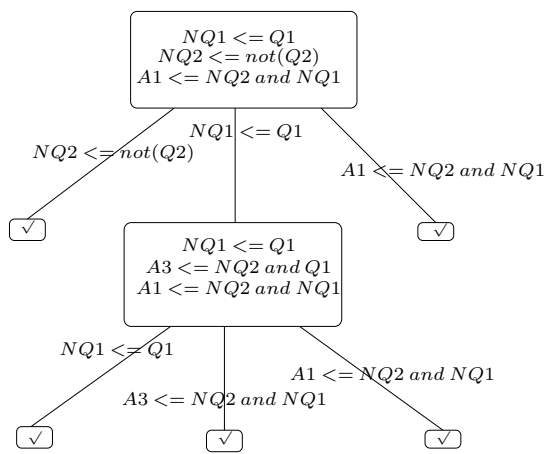}
\end{center}
\caption{The hitting set tree for the conflicts of the counter.}
\label{hs_tree}
\end{figure}

Figure \ref{hs_tree} outlines the hitting set tree for the conflicts we
obtained from our example. By collecting the edge-labels from the leaves
marked with $\surd$ up to the root in a focused breath-first search order 
all minimal diagnoses can be retrieved from the hitting-set tree
efficiently.

For example, considering the first level of the tree we obtain
the statements \{$A1<=NQ2$ $and$ $NQ1$\} and \{$NQ1<=Q1$\} (the introduced bug) as
single-fault diagnoses. Furthermore, taking into account the next level of the hitting-set
tree, there is a single dual-fault diagnosis : $\{NQ2<=NOT(Q2), A3<=NQ2$ $and$
$Q1$\}. The remaining diagnoses at this level can be ruled out; they do not
represent minimal diagnosis because a proper subset is already known as a
single-fault diagnosis. Usually, we prefer single-fault diagnoses, since they
are more probable. However, note that although \{$A1 <= NQ2$\} is the
real cause of the misbehavior, a correct behavior for the given test case can
also be obtained by appropriately modifying all statements of one of the 
the remaining diagnoses.

\section{Using Temporal Abstraction For Fault Localization}     

\label{temporal_abstraction}

In the following we present a model extension that allows not only for localizing
a fault at a given point in time, but also exploits the temporal
behavior of a circuit. The model extension is put on top of the model
introduced previously and allows for representing state explicitly but still
can be used with a standard diagnosis engine. As in the previous section, we first
introduce the structure of the new model and afterwards outline the behavioral
description of the components. When dealing with temporal aspects of VHDL-RTL
designs it is reasonable to require some restrictions.

For our research prototype we required that signal-assignment statements do
not contain a VHDL after clause. Processes are required to have an explicit
sensitivity-list and are not allowed to contain wait statements. Furthermore,
we require the process activation graph to be acyclic. Thus, considering a VHDL
program that comprises $n$ processes each process can at most be activated by
$n-1$ other processes. Taking into account the initial activation, there
is an upper bound of $n$ process activations for a single process per
simulation cycle. 

\begin{figure}
\begin{center}
\input{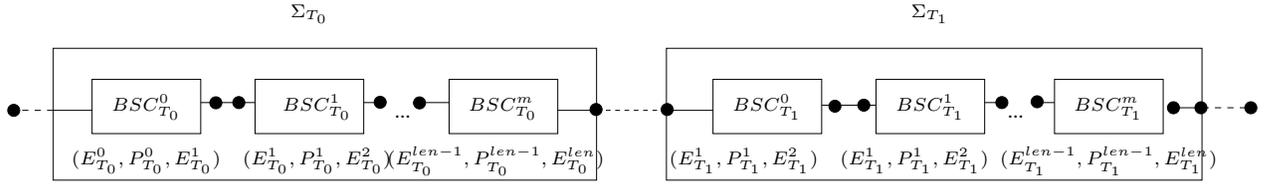}
\end{center}
\caption{The structure of the model in terms of the introduced notations.}
\label{struct_temporal_model}
\end{figure}

Simulating a program corresponds to computing all the values of the signals
and variables for a ordered set of times. At a technical level, the
simulation of a program is decomposed into atomic steps referred to as basic
simulation cycles (BSCs). A BSC at a specific point in time
$T$  and a given state $I$ of a VHDL program executes all processes for which
at least one signal on which an event occurred, appears in the sensitivity-list, and updates
the signal values according to the semantics of VHDL. Formally, a BSC can be
expressed as a triple $(E_T^I, P_T^I, E_T^{I+1})$ where $E_T^I$ denotes the
input environment and $E_T^{I+1}$ stands for the bindings of signals and
variables to their values after executing the processes in $P_T^I$. Figure
\ref{struct_temporal_model} outlines the way in which BSCs are connected. Note that the output
environment of a certain BSC is the input environment of the succeeding one.

Furthermore, the figure outlines a so called simulation sequence, the definition
of which is given in the following. A simulation sequence $\Sigma_T$ at time $T$ of a RTL program is a
sequence of basic simulation cycles $BSC^0_T, BSC^1_T, \ldots,
BSC^{len}_T$, where $len$ denotes the length of $\Sigma_T$. The state number of the last element of the simulation
sequence is called the length of the simulation sequence $len$.
In order to state the treatment of variables and signals between successive
simulation sequences we have to introduce an interpretation function $E_T^I:
SIG \cup VAR \mapsto VAL$ that
maps signals $SIG$ and variables $VAR$ to their corresponding values
$VAL$. By using this function the initial input values of a simulation sequence
$\Sigma_{T_j}$ can be stated as follows. 

\begin{center}
$E^0_{T_j}(X) = \left\{\begin{array}{ll}
E^0_{T_j}(X)   &  \mbox{ j = 0 $\vee X \in dom(E_{T_j}^0)$} \\
E^{len}_{T_{j-1}}(X) & \mbox{ otherwise} \\
\end{array} \right.$
\end{center}

In the formula given above $X$ denotes a variable or signal and the function
$dom$ retrieves the names of the signals and variables of a given environment.
The formula states that the state at time
point $T_{j-1}$ is used for performing the simulation at time point $T_j$ except from
those variables and signals that are explicitly given in the input environment
$E_{T_j}^0$. Note that the formalization requires an initial binding for every
signal or variable $X$ in $E_{T_0}^0$. This approach allows for representing so
called external signals, the values of which are given from the outside
(e.g. a $CLK$ signal), in our model.

By using the terms and notations introduced above we are able to describe 
our diagnosis model in terms of BSCs. Formally, the system
description for a simulation sequence $S_{T_0}, S_{T_1}, ...,
S_{T_k}$ where each $S_{T_i}$ is a sequence of BSCs $BSC_{T_i}^0, ...,
BSC_{T_i}^{len}$ is given by the following logical sentence.

\begin{center}
$SD = \bigcup_T SD(S_T) \cup CONN, SD(S_T) = \bigcup_{i=1}^{len(S_T)} SD(BSC_T^i)$
\end{center}

$SD(BSC_T^i)$ is composed from the system description at a single point in
time $T$ and also covers the event handling according to the VHDL
semantics. Moreover, $CONN$ denotes logical sentences that describe the
connection between the BSCs itself (depicted by solid lines in Figure
\ref{struct_temporal_model}) as well as BSC sequences (the corresponding
connections are outlined by using a dashed line). These connectivities 
are given by the signals that a temporal process instance uses likewise as input and
output. Using this approach our model can handle state by unfolding process executions
over time. The handling of events and the logical description of a process
component remains to be introduced. 

VHDL designs consist of processes that are assumed to be executed in
parallel and which communicate by means of signals. A process $p$ is
executed if at least one of the signals that occur on its sensitivity-list has
changed its value in the preceeding simulation cycle. If this is the case, the
values of the signals used as a target in the sequential statement part are
computed according to the VHDL semantics. If none of the signals
within the sensitivity list has changed its value, then the original input
values before executing the sequential statement block of process $p$ are
propagated to the output of the process component. In the following a logical
description of a process component $p$ is outlined:

\begin{center}
$ \forall y \in inputs(p) \cdot \exists x \in sensitivity-list(p) \cdot$ \\
\hspace*{1cm} $ s(x) = true \rightarrow out_{y}(p) = in_{y}(p) $ \\
$\forall y \in inputs(p) \cdot \forall x \in sensitivity-list(p) \cdot$ \\ 
\hspace*{1cm} $s(x) = false \rightarrow out_{y}(p) = def_{y}(p) $ \\
$out_{y}(p) = def_{y}(p) \leftrightarrow out_{y'EVENT}(p) = false$ \\
$out_{y}(p) \neq def_{y}(p) \leftrightarrow out_{y'EVENT}(p) = true$
\end{center}

In the logical sentences above $inputs(p)$ denotes the set of inputs of the process $p$ and
$sensitivity-list(p)$ represents the sensitivity list of the process
$p$.  The predicate $s(x)$ is set true if signal $x$ has changed its value. The
signal value $y$ after executing the sequential statement part of process $p$ is
represented by $in_{y}(p)$ whereas the corresponding unmodified input
of the process $p$ is represented in formal terms by
$def_{y}(p)$.  Furthermore, if an output has changed, the corresponding event
signal is set to true.

Beside the system description $SD$ we have to define the observations
$OBS$ and the set of components $COMP$ before we are able to use the
model-based diagnosis approach. The set of observations $OBS$ for our
model is equal to the test stimulus, that is the sequence of test cases 
of the original program. The set
of components is given by the set of temporal component instances, i.e.,
$COMP^*=\bigcup_{i=1}^{len(S_T)}\{Stmt(p_T^i) | p_T^i \in BSC_T^i \}$ for every
simulation cycle $S_T$ where $Stmt$ returns all
diagnosis components that correspond to the process instance. 
Diagnosis components that represent the same statement in the code but belong
to different process instances are treated as different components. A multiple-fault diagnosis thus may exclusively contain components corresponding to the
same statement in the source code. 

The model extension discussed above improves the existing model with respect
to two aspects. First, it is closer to the semantics of VHDL taking into
account the treatment of processes and events on signals. Second,
the extension handles the temporal behavior of VHDL by unfolding the process
executions over time. Hence, unlike the previous models, our model introduces
temporal aspects of diagnostic reasoning about VHDL programs.
 
However, since temporal instances of a components are handled completely
different, the number of components that may account for a certain
misbehavior may increase in comparison to the original model. Thus,
the obtained diagnoses are mapped back by
reducing the instances to their corresponding components in accordance
to the following procedure.
\begin{center}
$
  \begin{array}{l}
    \varphi : \Delta^* \subseteq COMP^* \mapsto \Delta \subseteq COMP \\
    \varphi(\Delta^*) = \{C_i | C_i^{(j)} \in \Delta^*\} \\
  \end{array}
$
\end{center}
In the mapping given above $COMP^*$ denotes the instances of the
components whereas $COMP$ refers directly to the components. In
similar fashion, $\Delta^*$ stands for the diagnoses that are computed
using temporal instances and $\Delta$ simply denotes the corresponding
components. Moreover, $C_{i}^{(j)}$ denotes the instance $j$ of
component $C_{i}$.

\section{Empirical Results}

\label{empirical_results}

\begin{figure}
\begin{center}
\includegraphics [width = \linewidth] {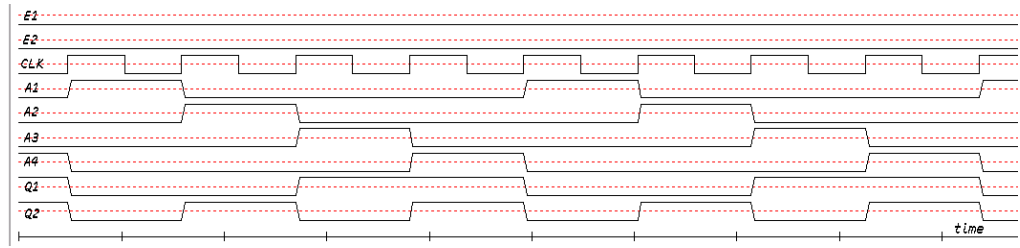}
\end{center}
\caption{The wave trace of the correct program.}
\label{correct_wavetrace}
\end{figure}

\begin{figure}
\begin{center}
\includegraphics [width = \linewidth ]{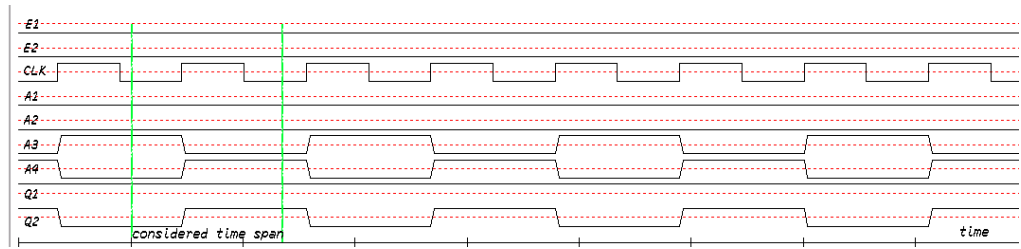}
\caption{The wave trace of the faulty VHDL design .}
\label{faulty_wavetrace}
\end{center}
\end{figure}

\begin{figure}
\begin{center}
\input{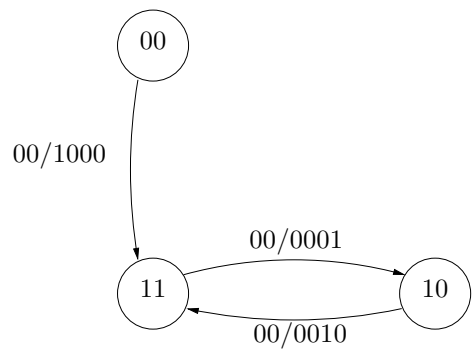}
\end{center}
\caption{The state-transitions for the faulty program when counting upwards.}
\label{state_transition_and}
\end{figure}

In order to demonstrate the applicability of the new model we briefly outline
a fault localization scenario. Therefore we introduce a bug
in line 13 by substituting the {\tt or} operator by an {\tt and} operator:

\scriptsize{
\begin{alltt}
    comb_in : process (Q1, Q2, E1, E2)
         variable I1, I2 : bit;
       begin
         I1 := not((Q1 and Q2) and (not(Q1) and not(Q2)));
         I2 := (I1 and E1) or (not(I1) and not(E1));
         ...
         ...
\end{alltt}
}
\normalsize
Figure \ref{correct_wavetrace} outlines the expected behavior in the time
domain and Figure \ref{faulty_wavetrace} shows the observed wave trace and
also indicates the time span taken into account for computing
diagnoses. The state-transition diagram in Figure \ref{state_transition_and}
shows the faulty behavior encountered when counting upwards.
Since we assumed the process activation graph to be acyclic, for
every process we have to reserve at most 5 instances for a single point in
time. In total, since we considered 3
points in time, 15 instances have to be forseen for unfolding a certain
process in time. Table \ref{tab_obs}
collects the observations, where $t_0$ denotes the time at which diagnosis is
started and $t_2$ refers to the point in time at which the expected values are
specified, a '-' indicates that the signal is not used as an observation.

\begin{table}[h]
\begin{center}
  \scriptsize{
\begin{tabular}[h]     
{|    c   |ccc|}

\hline
                   &
                  $t_0$ &
                  $t_1$    &
                  $t_2$ \\

                  &  &   &  \\ 

\hline
 CLK & 0 & 1 & 0\\
 A1 & 0 & - & 0\\
 A2 & 0 & - & 0\\
 A3 & 1 & - & 0 \\
 A4 & 0 & - & 1 \\
\hline
\end{tabular}
}
\end{center}
\label{tab_obs}
\caption{The observations that are provided for fault localization.}
\end{table}

After invoking the diagnosis procedure we obtained 25 single-fault diagnoses and 66
dual-fault diagnoses where each dual-fault diagnosis can be mapped back to a single
bug in the source code when using the $\varphi$ function. In summary the diagnosis
correspond to the statements  10, 13, 14, 15, 19, 38, 40 and 41 and include
the introduced bug. The statements 10, 19, 38 and 40 correspond to a process
or a conditional statements. From the remaining statements 13, 14, 15 and 41
the lines 13, 14 and 15 correspond to functional faults,
i.e., wrong operators in the code. This result indicates that the new approach
can localize the statement that is responsible for the misbehavior. 

The new approach has been evaluated using 3 different counters. 
We observed that, although the quality of the computed diagnosis may
vary depending on the example being used and the observations taken into
account, the faulty component is always present in the computed
diagnoses. Note that any of the obtained diagnoses explains the faulty
behavior with respect to the given test case. However, the main obstacle towards an industrial-sized application of
our approach is the computational effort that is necessary for localizing faults in
larger sized real-world examples, particularly when several intermediate
states are to be considered and dual- or tripe-fault diagnoses have to be
computed. This problem not only arises in the context of temporal unfolding,
even when considering real word sized combinatorial circuits we observed that
depending on the structure of the circuit and the specific test case
being used there is a large variation in the elapsed running time for
computing all single-fault diagnoses. 

The ISCAS'85 benchmark is a suite of combinatorial circuits used by many
researchers in the field of verification, automatic test pattern generation and diagnosis. By using
this suite of circuits we evaluated the running-times that have to be expected
when dealing with real world problems. We have run 100 test vectors for every circuit where the test
 cases have been created in such a way that at least one single-fault diagnosis can be
 obtained for every test case. Each computation has been repeated 10 times and the minimum,
 the maximum, the average and the median value of the elapsed running time excluding
 model-setup time has been recorded. The running-times include the
 time for
 storing all the diagnoses being found. In addition, the average number of single-fault
 diagnoses as well as the median value of the number of single-fault diagnoses have
 been recorded. Table \ref{fig_diags} outlines the results with respect to the number of
 single-fault diagnoses that have been obtained and Table \ref{fig_runningtimes} collects the
 elapsed running-times for computing all single-fault diagnosis.
The computations were carried out on a slightly loaded Pentium 4/1.8 Ghz
machine using a Visual-Works Non-Commercial Smalltalk programming
environment in version 5i.4.

\begin{table}[h]
  \begin{center}
    \scriptsize{     
 \begin{tabular}{|c|ccccc|}
        \hline
        circuit name &
        median nr.   &
        avg. nr.  &
        gates &
        inputs &
        outputs \\      
  
        & (single-fault diag.) & (single-fault diag.) &  &  &  \\ 
        \hline
        C432 & 1 & 1.40 & 160 & 32 & 7 \\
        C499  & 2 & 2.43 & 202 & 41 & 32 \\
        C880 & 4 & 4.06 & 383 & 60 & 26\\
        C1355 & 5 & 5.53 & 546 & 41 & 32 \\
        C1908 & 31 & 23.78 & 880 & 33 & 25  \\
        C2670 & 2 &  2.78 & 1193 & 233 & 140 \\
        C3540 & 4 &  5.67 & 1669 & 50 & 22 \\
        C5315 & 3 &  3.12 & 2307 & 178 & 123 \\
        C6288 & 306 & 287.02 & 2406 & 32 & 32 \\
        C7552 & 1 & 4.75 & 3512 & 207 & 108 \\
        \hline
        
      \end{tabular}
      }
    \caption{Average and median number of diagnosis.}
    \label{fig_diags}
  \end{center}
\end{table}
\begin{table}[h]
  \begin{center}
    \scriptsize{     
 \begin{tabular}{|c|cccc|}
        \hline
        circ. name &
        min run times &
        median run times  &
        avg run times &
        max run times \\      
  
        & [ms] & [ms] & [ms]  & [ms]   \\ 
        \hline
        C432 & 3 & 14 & 14 & 41 \\
        C499  & 8 & 28 & 29 & 109 \\
        C880 & $<$ 1 & 27 & 25 & 133 \\
        C1355 & 80 & 254 & 273 & 943 \\
        C1908 & 121 & 1397 & 1411 & 4691  \\
        C2670 & $<$1 & 116 & 170 & 2292  \\
        C3540 & 377 & 2267 & 3140 & 17352  \\
        C5315 & $<$ 1 & 492 & 520 & 3287  \\
        C6288 & 3926 & 122317 & 138910 & 2792591 \\
        C7552 & $<$1 & 958 & 8219 & 75735 \\
        \hline
      \end{tabular}
      }
    \caption{Recorded running times for the different circuits.}
    \label{fig_runningtimes}
  \end{center}
\end{table}

\section{Related Work, Conclusion and Further Research}

\label{conclusion}

Shapiro presented methods and algorithms for fault
localization in logic programs \cite{sha83}, commonly referred to as
algorithmic debugging. In \cite{con93} some advantages of the model-based approach over
the algorithmic debugging approach are pointed out. However, as
outlined in \cite{bond96}, the model-based approach and algorithmic
debugging are part of the same spectrum provided a logic program is
used as the system description. Kuchcinski et al \cite{kuch93} discusses an
application of algorithmic debugging to automatic fault localization
in VLSI designs and proposes a method for
smooth combination of different diagnosis techniques, where the use of
logic specifications and algorithmic debugging plays an essential
role. The authors of \cite{kuch93} summerize, that algorithmic
debugging is similar to model-based diagnosis in that it uses a model
of the correct expected behavior of the program. However, in algorithmic
debugging the model is
used in a probing-controlled search for faulty components rather than
for generation of an exhaustive set of diagnosis hypotheses. The
relationship of model-based diagnosis and algorithmic debugging is
discussed in \cite{con93}.

In this paper we show how to create models for software debugging of hardware
designs. The paper in detail discusses our value-level model for VHDL-RTL
designs and shows the computation of fault locations by means of a small but
practical example. Furthermore, a new model is introduced that allows for
localizing faults considering the signal values during a whole time span rather
than that of a certain point in time. This can be reached by unfolding our
original value-based model in time and incorporating it with the semantics of process
executions and event handling. Furthermore, an example indicates the principal
applicability of the new approach. Moreover, the paper outlines some
empirical results on the computational effort that currently is required in
software debugging of real-world designs.

Besides from improving scalability, further research shall deal with model improvements by means of forward
simulation, i.e., exploiting the results of a simulation to further improve
our model. Recently such an approach was successfully applied in automated
debugging of Java programs where the behavioral description of an if-then-else
statement was considerably improved \cite{sww01b}. Another direction of current research is
a detailed analysis of the outcome of our approach. Such analysis should give
answers to the question whether there are kinds of faults that cannot be
detected. 

Further topic of future research is to incorporate the domain of verification and debugging.
Verification techniques require a separate specification that, however, is
often not available. If one is available though, techniques such as model
checking can produce counterexamples for violated properties. Using these
counterexamples to generate input for diagnosis is also part of
future research. 

\section{Acknowledgments}

The work was partially supported by the Austrian Science Fund (FWF)
under project grant P15163-INF.

\bibliography{aadebug}

\end{document}